\documentclass{article}

\usepackage{arxiv}

\usepackage[utf8]{inputenc} 
\usepackage[T1]{fontenc}    
\usepackage{hyperref}       
\usepackage{url}            
\usepackage{booktabs}       
\usepackage{amsfonts}       
\usepackage{nicefrac}       
\usepackage{microtype}      
\usepackage{lipsum}
\usepackage{graphicx}
\usepackage{amsmath}
\usepackage{amssymb}
\usepackage{multirow}
\graphicspath{ {./images/} }
\newcommand{\tabincell}[2]{\begin{tabular}{@{}#1@{}}#2\end{tabular}} 
\newcommand{\printfnsymbol}[1]{%
  \textsuperscript{\@fnsymbol{#1}}%
}
\title{Enhancing Social Relation Inference with Concise Interaction Graph and Discriminative Scene Representation}

\author{
 Xiaotian Yu\thanks{equal contribution} \\
  Intellifusion\\
  \texttt{yu.xiaotian@intellif.com} \\
   \And
 Hanling Yi\footnotemark[1] \\
  Intellifusion\\
  \texttt{yi.hanling@intellif.com} \\
  \And
 Yi Yu \\
  Intellifusion\\
  \texttt{yu.yi@intellif.com} \\
  \And
 Ling Xing \\
 Intellifusion \\
 \texttt{xing.ling@intellif.com} \\
  \And
  Shiliang Zhang \\
  Peking University \\
  \texttt{slzhang.jdl@puk.edu.cn} \\
  \And
  Xiaoyu Wang \\
  Intellifusion \\
  \texttt{wang.xiaoyu@intellif.com}
}

\begin{document}
\maketitle
\begin{abstract}
There has been a recent surge of research interest in attacking the problem of social relation inference based on images. Existing works classify social relations mainly by creating complicated graphs of human interactions, or learning the foreground and/or background information of persons and objects, but ignore holistic scene context. The holistic scene refers to the functionality of a place in images, such as dinning room, playground and office.
In this paper, by mimicking human understanding on images, we propose an approach of \textbf{PR}actical \textbf{I}nference in \textbf{S}ocial r\textbf{E}lation (PRISE), which concisely learns interactive features of persons and discriminative features of holistic scenes. Technically, we develop a simple and fast relational graph convolutional network to capture interactive features of all persons in one image. To learn the holistic scene feature, we elaborately design a contrastive learning task based on image scene classification. To further boost the performance in social relation inference, we collect and distribute a new large-scale dataset, which consists of about 240 thousand unlabeled images.  The extensive experimental results show that our novel learning framework significantly beats the state-of-the-art methods, e.g., PRISE achieves 6.8$\%$ improvement for domain classification in PIPA dataset.
\end{abstract}


\section{Introduction}
Social relations describe the connections among two or more individuals, which are fundamental to daily life of human beings~\cite{kitayama2000pursuit}.  Nowadays, billions of people share images in social media platforms such as Facebook and Twitter. In light of~\cite{bromley1988understanding}, common social relations include family, couple, friends, colleagues, professional, etc. There has been an increasing interest in understanding social relations among persons from still images due to the broad applications including group behavior analysis~\cite{hoai2014talking}, image caption generation~\cite{khademi2018image} and human trajectory prediction~\cite{alahi2016social}.

\begin{figure}[t]
\centering
\includegraphics[width=3.3in]{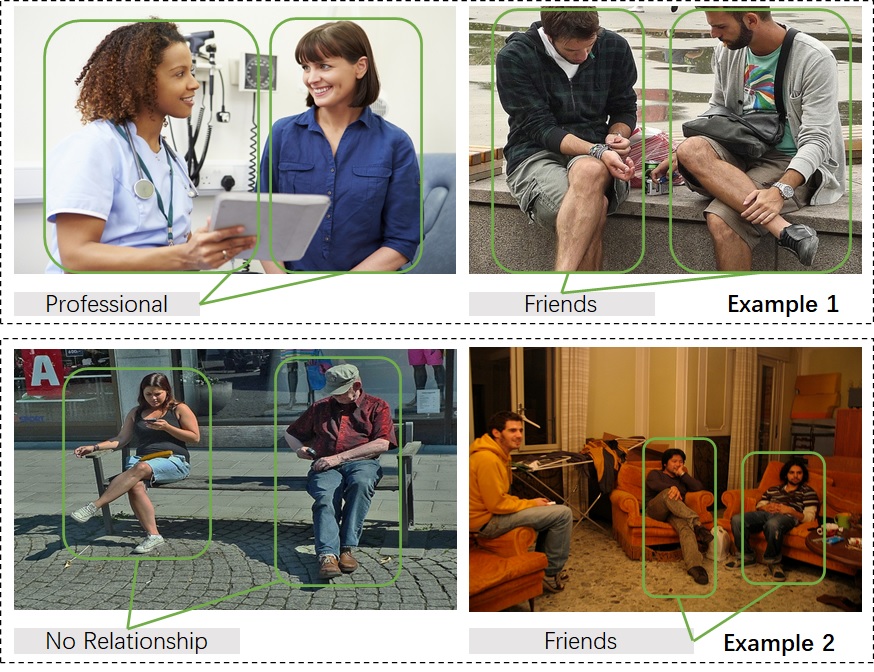}
\caption{The comparison of inferring social relations under different scenes, with images taken from PISC dataset~\cite{li2017dual}. Example 1 shows relations of professional and friends corresponding to hospital and park, respectively. Example 2 significantly implies a close relation in the context of staying indoors.}
\label{fig:comparison_of_scene}
\end{figure}

The problem of social relation inference is challenging and complicated because it requires high-level semantic understanding of images. Inspired by the cognition process of human beings on images, we summarize three steps for classification of social relations. First, we take a whole view of the scene, which represents the functionality of a place. Second, we identify the background objects and persons, and the foreground union regions of person pair in images. Third, we observe the interaction of persons, such as hugging and handshaking. With these information in mind, we infer the category of relationships for all persons. 

To demonstrate the importance of scene at inferring social relations, we present two examples in Figure~\ref{fig:comparison_of_scene}. Each example consisting of two images shows different relationships mainly due to the scene context. For instance, Example 1 shows the professional relationship in the context of hospital and the relationship of friends in the context of a park. It is clear that the scene information should be carefully taken into consideration for social relation inference.


We note that prior works put more effort into learning from human interactions, the foreground union regions of person pair, and the background information of persons and objects, but missed the importance of holistic scene context, especially in inferring relationships from different scenes with similar human interactions. Goel et al.~\cite{goel2019end} adopted a pre-trained model to directly output the feature of foreground regions, which missed the whole context and may get a less representative feature. Zhang et al.~\cite{zhang2019multi} utilized a pre-trained model with ImageNet to generate a feature at the level of object classification. The resultant feature cannot semantically summarize high-level scene features.

A large number of studies proposed models to learn social relations based on interaction graph of persons in an image~\cite{goel2019end,li2020graph,wang2018deep,zhang2019multi}. Wang et al.~\cite{wang2018deep} proposed graphs of persons and objects to infer social relations. The significant drawback of~\cite{wang2018deep} is that graphs of persons and objects can only characterize the connection between two persons, which leads to complicated calculations for cases of three or more persons in an image.

It is urgent and essential to attack the problem of social relation inference in a broader view where an interaction graph works for two or more persons and simultaneously the scene feature provides holistic hints for the classification of relationships. In this paper, inspired by the understanding process of human beings, we propose an approach of \textbf{PR}actical \textbf{I}nference in \textbf{S}ocial r\textbf{E}lation (PRISE), which synthesizes three streams of information, i.e., holistic scenes, foreground and background information of persons and objects, and interaction of persons. To the best of our knowledge, we are the first to methodically and systematically develop a model to learn the holistic scene feature in social relation inference based on contrastive learning.



In PRISE, we first technically design a concise relational graph convolutional network (RGCN) to extract the interactive features of all persons in one image. Then, to boost the performance in social relation inference, a contrastive learning task for capturing holistic scene is incorporated into the proposed PRISE. Intuitively, the contrastive learning task helps to extract discriminative features of the holistic scene context. We demonstrate that PRISE achieves a significant improvement in social relation inference compared to the state-of-the-art methods.



We summarize our contributions in this work as follows.

\begin{itemize}
    \item We systematically develop a novel approach, i.e., PRISE, for social relation classifications. PRISE significantly beats the state-of-the-art methods in social relation inference. 

    \item We design a concise relational graph convolutional network to capture the interactive features for all persons. The proposed RGCN is simpler and faster than the graph model in~\cite{li2020graph}.

    \item We construct a contrastive learning task to learn discriminative representation of holistic scene. We distribute a new large-scale dataset for contrastive learning, which is named as PISC-extension. The usefulness of PISC-extension can be extended to other tasks in computer vision, such as group behaviour analysis.

    \item Extensive experiments including a comprehensive ablation study demonstrate the effectiveness the PRISE, and show the significance of interaction graph and scene information in social relation inference.
\end{itemize}

\section{Related Work}
To assess our contributions in classification of social relations, it is important to consider three streams of studies: social relation inference, graph neural networks and contrastive learning. 

\subsection{Social Relation Inference}
For a large number of scenarios in computer vision, social information has played an important role by providing additional cues in tasks of image understanding, e.g., human interaction~\cite{shu2019hierarchical}, kinship recognition~\cite{lu2017discriminative,liang2018weighted,lu2013neighborhood} and image caption generation~\cite{xu2015show}. 

The pioneering work on social relation inference dates back to 2010 from~\cite{wang2010seeing}, where the authors developed a model to characterize the interaction between multi-person actions, facial appearances and identities. 
Zhang et al.~\cite{zhang2015learning} developed a deep neural network to learn social relation traits from rich facial attributes, such as expression, gender, and age. In~\cite{zhang2015learning}, the social relation traits were defined based on psychological studies~\cite{hess2000influence,gottman2001facial}, consisting of eight types, e.g., trusting and friendly.

For datasets in social relations, Zhang et al.~\cite{zhang2015beyond} distributed a dataset to evaluate classification of social relations, which is named as People In Photo Albums (PIPA). Besides, another dataset, which is People in Social Context (PISC), was published in~\cite{li2017dual}. 

With PIPA and PISC, several interesting works move forward along the research line of social relation understanding~\cite{sun2017domain,wang2018deep,goel2019end,li2020graph}. In light of domain based theory from social psychology, Sun et al.~\cite{sun2017domain} presented a model with semantic attributes to classify social relations and domains. Wang et al.~\cite{wang2010seeing} modelled a knowledge graph with proper messages propagation and attention to learn the social relations among people in an image. Recently, in~\cite{goel2019end}, Goel et al. proposed an end-to-end neural network to learn the interaction graph of persons. In~\cite{li2020graph}, a social graph was proposed to restrict logical connections of persons, which achieved the state-of-the-art results in social relation inference. In Table~\ref{tab:com.feature}, we present the differences between our PRISE and the prior studies in terms of feature information.
\begin{table}[t]
    \centering
    \caption{Comparisons between our PRISE and previous methods in features for social relation inference. ``Fore." is short for ``Foreground" and ``Back." is short for ``Background".}
    \begin{tabular}{l|c|c|c}
    \toprule
        Methods & \tabincell{c}{Interaction \\Graph} & \tabincell{c}{Fore. \&\\ Back.} & \tabincell{c}{Holistic \\ Scene} \\
        \midrule
        Pair CNN~\cite{li2017dual}  & No & Yes & No\\
        Dual-Glance~\cite{li2017dual} & No & Yes & No \\
        SRG-GN~\cite{goel2019end} & Yes & Yes & No\\
        GRM~\cite{wang2018deep} & Yes & Yes & No\\
        MGR~\cite{zhang2019multi} & Yes & Yes & No\\
        GR$^2$N~\cite{li2020graph} & Yes & No & No\\
        \midrule
        PRISE &  Yes & Yes & Yes\\
    \bottomrule
    \end{tabular}
    \label{tab:com.feature}
\end{table}

\subsection{Graph Neural Networks}
Inspired by the success of convolutional networks in the computer vision domain, GNNs are proposed to re-define the notation of convolution for graph structured data~\cite{kipf2016semi}.
Most recently, GNNs have been adopted to social relation reasoning~\cite{wang2018deep,zhang2019multi,li2020graph}. 
For instance, 
Zhang et al.~\cite{zhang2019multi} designed person-object graph and person-pose graph, and conducted social relation reasoning on these two graphs by GNN. Li et al.~\cite{li2020graph} proposed a graph relational reasoning network to jointly infer social relations by building a graph for each image, where the nodes represent the persons and the edges represent the relations. In this paper, we follow the similar graph-based approach proposed in~\cite{li2020graph}, and design a concise relational graph convolutional network to extract interactive features among people in the image.

\subsection{Contrastive Learning}

Over the last few years, contrastive representation learning based on deep learning models has shown the power in many practical tasks~\cite{chen2020simple,devlin2018bert,doersch2015unsupervised,pathak2016context,xiao2020should,yang2019xlnet,zhang2017split}, especially for natural language and computer vision domains. Contrastive learning usually maximizes similarity and dissimilarity over data samples which are organized into similar and dissimilar pairs, respectively. 

A significant challenge in contrastive learning is how to select the similar (or positive) and dissimilar (or negative) pairs. The main difference among different approaches of contrastive learning lies in their strategy for obtaining sample pairs~\cite{chuang2020debiased}. To generate sample pairs without additional human labels, many researchers create models with multiple views of each sample. Besides, for complicated tasks, some studies also construct positive and negative sample pools from pre-trained models~\cite{feng2019self,lee2019multi}. In this paper, by utilizing the pseudo-labels from a pre-trained model of image scene classification, we construct negative and positive sample pools, and design a contrastive learning task to learn discriminative scene representations.

\section{Methodology}

\begin{figure*}
\centering
 \includegraphics[width=0.95\textwidth]{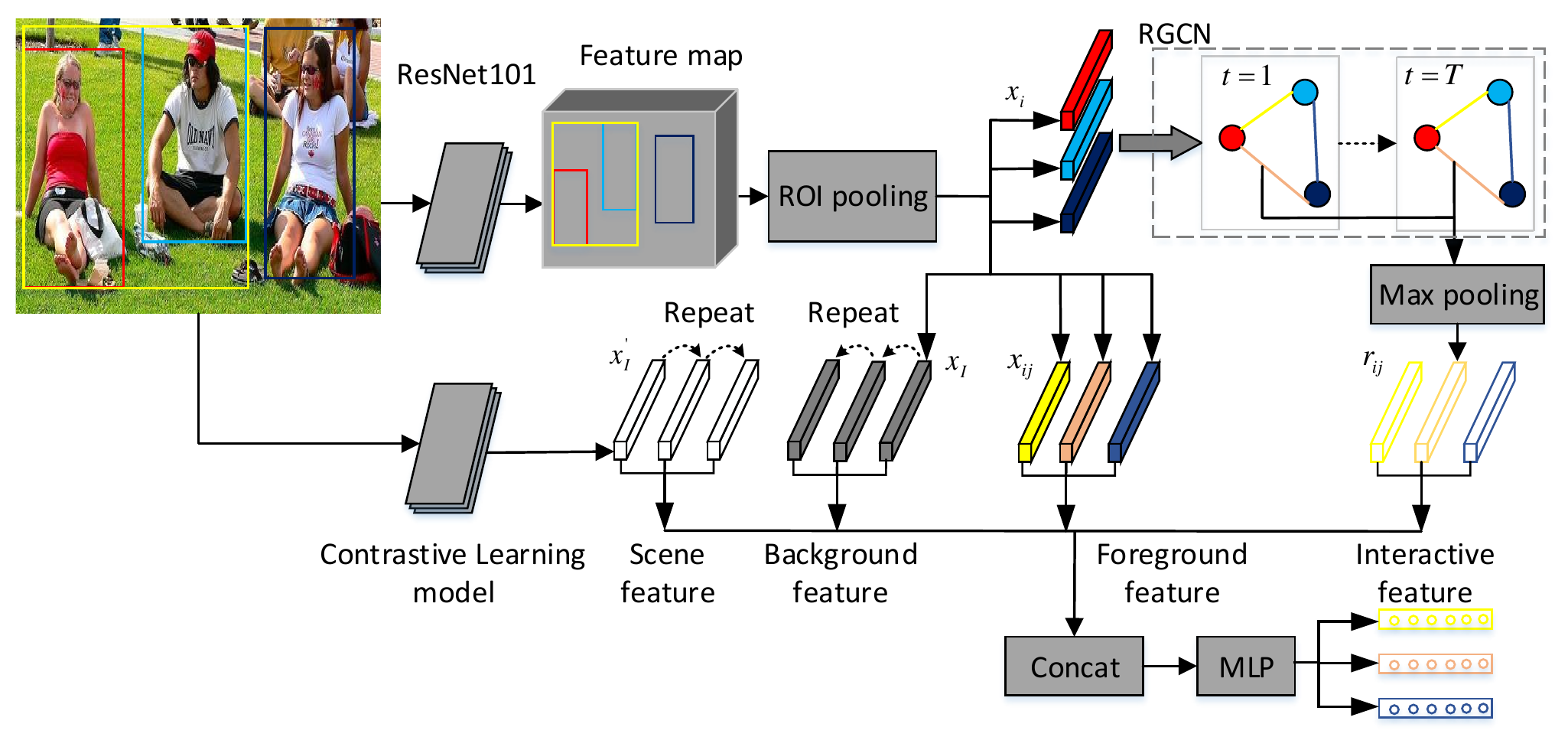}
 \caption{\label{fig:network}The overall pipeline of PRISE model. Given an input image $I$, we use ResNet101 to extract RoI features of people in the image $x_i$, foreground feature $x_{ij}$ and background feature $x_{I}$. In addition, another pre-trained ResNet50 that was finetuned using contrastive learning approach to extract discriminative scene feature $x'_{I}$. The RGCN is used to obtain interactive feature between person pair $r_{ij}$.
Finally, $r_{ij}, 
 x_{I}, x_{ij}, x'_{I}$ are concatenated and passed to a MLP layer for relation classification. The network outputs relational class distribution for all person pairs in the image. The operation `Repeat' is a must to keep the number of scene and background features the same as the number of person pairs when there exist more than two persons in an image.}
\end{figure*}

In this section, we first introduce the approach that converts all persons in an image into an interaction graph, and then apply the RGCN model on the graph to learn interactive features of people in the same image. Finally, to better utilize scene information for social relation understanding, we propose a contrastive learning approach to learn discriminative scene representation. The overall pipeline of PRISE is shown in Figure~\ref{fig:network}.

\subsection{Graph-based Approach}
\label{sec:graph-based_approach}
Inspired by~\cite{li2020graph}, we adopt graph-based approach. We build a graph for each image, where each person in an image is modeled as a node in the graph. The edge between two nodes represents the social relation between the corresponding two persons. For simplicity, we consider the fully connected graph, i.e., each pair of persons in the image has an edge. Denote $\mathcal{G}=(\mathcal{V}, \mathcal{\xi})$ as the fully connected graph with node set $\mathcal{V}$ and edge set $\mathcal{\xi}$ in an image.

For each image, we extract three types of features using a ImageNet pre-trained model (i.e., ResNet101). These three types of features include RoI features of single person, union region of person pairs (a.k.a. foreground feature), and persons and objects of the whole image (a.k.a background feature). In the following, we will introduce the detailed ways to generate these features.

Following traditional approach in detection, the feature representation of each person is extracted directly from the last convolutional feature map of the input image. Specifically, given input image $I$ with $N$ bounding boxes $b_1, b_2, ..., b_N$ for $N$ persons, we obtain the feature representations of all people in the image using a pre-trained ResNet101 model, where an RoI pooling layer is constructed based on the last convolutional feature map. Note that the RoI pooling layer is a common trick in social relation learning with graph representation~\cite{li2020graph}. Denote the feature representation of the $i$-th person in image $I$ as $x_i$, 
\begin{equation}
    x_i = f_{CNN-RoI}(I, b_i)\in \mathbb{R}^F, i=1,2,...,N,
\end{equation}
where $F=2048$ is the feature dimension for each person. For simplicity, we denote the set of feature representations for people in image $I$ as $X=\{x_1, x_2, ..., x_N\}$.

In addition, we obtain the features of union regions of person pair using the same approach. For person $i$ and $j$, we first compute the bounding box of their union region $b_{ij}$. Then we get its feature as follows:
\begin{equation}
    x_{ij} = f_{CNN-RoI}(I, b_{ij}).
\end{equation}
Besides, we also obtain $x_{I}$, the feature representation for the whole image, by setting the bounding box to cover the whole image and passing it to $f_{CNN-RoI}$.
\begin{equation}
    x_{I} = f_{CNN-RoI}(I, b_{I}),
\end{equation}
where $b_{I}$ is the bounding box for the whole image. Intuitively,  the feature of single person $x_i$ encodes personalized
information of each person, the feature of union region $x_{ij}$ encodes the pair-wise foreground information, while the feature of whole image $x_{I}$ encodes the background of all persons and objects. Thus, all these features can provide useful information for social relation understanding. 

\subsection{Relational Graph Convolutional Network}
In this section, we introduce RGCN, an end-to-end trainable network architecture, that can learn pair-wise interactive features given arbitrary graph structured data. We apply RGCN on the fully connected graph $\mathcal{G}$ with features $X$.

Given $\mathcal{G}$ and $X$, for each node $i\in \mathcal{V}$, we set its initial node feature vectors as $h_i^0=w x_i\in \mathbb{R}^F,\forall i\in \mathcal{V}$, where $w\in \mathbb{R}^{F\times F}$ is a learnable parameter that maps input feature vectors to the new feature space. Correspondingly, each edge has a feature vector, and we denote the initial edge feature vector between node $i$ and node $j$ as $r^0_{ij}\in \mathcal{\xi}$. In RGCN with $T$ layer, the edge and node feature vectors are updated iteratively for $T$ times. Specifically, at $t$-th layer the edge and node representations can be expressed as follows:
\begin{align}
    r_{ij}^t &= \sigma(W^th_i^t+W^th_j^t) \label{eq:edge_update}, \\
    h_i^{t+1} &=  h_i^t + \sigma(W^th_i^t+\sum_{j\in\mathcal{N}_i}r_{ij}^t\odot W^th_j^t) \label{eq:node_update},
\end{align}
where $\mathcal{N}_i$ is the set of neighbors for node $i$, $W^t\in \mathbb{R}^{F\times F}, t=1,2,...,T$ are the learnable parameters at each layer,  and $\sigma(\cdot)$ is the ReLU function. 

We note that the RGCN defined in \eqref{eq:edge_update}-\eqref{eq:node_update} is an anisotropic variant of GCN~\cite{dwivedi12benchmarking}. Similar to Residual GateGCN~\cite{bresson2017residual}, our RGCN has residual connections on the node feature representations, and explicitly maintains edge feature at each layer. Intuitively, the edge feature representations at different layers encode the pair-wise human interaction information. Following similar ideas in JK-Net~\cite{xu2018representation}, we obtain the final interactive features by using a max pooling on the edge representations from different RGCN layers. Formally, the final interactive feature between person $i$ and $j$, denoted as $r_{ij}$, can be expressed as  
\begin{equation}
    r_{ij} = f_{max}(r_{ij}^0, r_{ij}^1, ..., r_{ij}^T),
\end{equation}
where $f_{max}(\cdot)$ is an element-wise max function.

\subsection{Discriminative Scene Representation Learning}
The scene of an image provides important visual clues for social relation understanding.  For instance, given a party scene, the group of people are more likely to be friends than colleagues, and a group of athletes running on a track are much more likely to be sports team
members than band members~\cite{goel2019end}. 
To harvest the power of pre-trained CNN model and unlimited amount of unlabeled images, in this paper, we propose a contrastive learning (CL) approach for discriminative scene representation learning.

\begin{figure}
 \begin{center}
 \includegraphics[width=0.95\columnwidth]{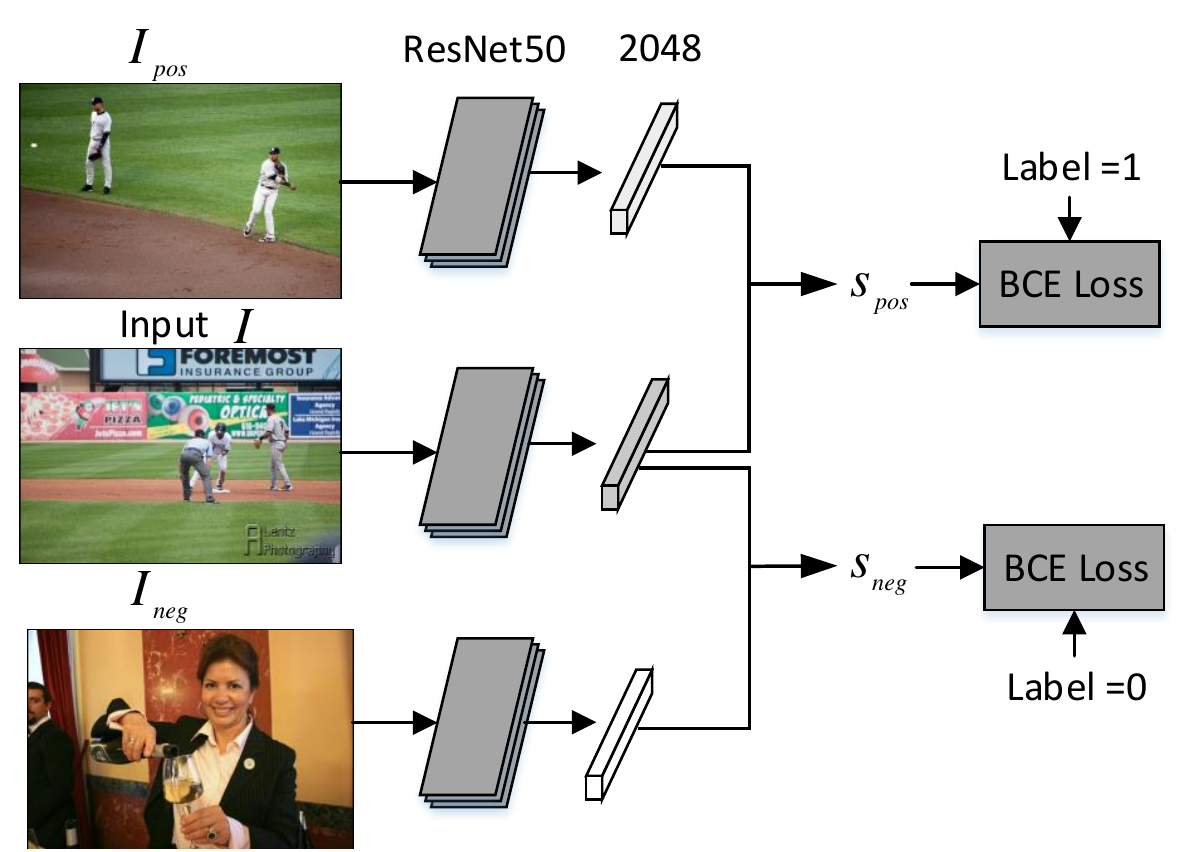}
 \end{center}
 \caption{\label{fig:ssl_pretext_task}An overview structure of CL task. For a given image $I$, we first sample a similar image $I_{pos}$ and a dissimilar image $I_{neg}$ from the image dataset. All these three images are passed through the pre-trained ResNet50 model to obtain a feature representation.}
\end{figure}


Following the CL paradigm, we design a scene classification task to distinguish between similar and dissimilar images. As a pre-process step, we use the pre-trained ResNet50 model~\cite{zhou2017places} to obtain the top-5 scene classes for each unlabeled image. Two images are defined as similar in scene if there are more than $K$ scene classes that are the same among the top-5 scene classes. Otherwise, they are dissimilar. Based on this definition, we can have for each image a pool of similar images and a pool of dissimilar images. The structure of CL task is shown in Figure~\ref{fig:ssl_pretext_task}.  For each input image $I$, we randomly sample one image $I_{pos}$ from its pool of similar images to construct positive sample, and another image $I_{neg}$ from its pool of dissimilar images to construct negative sample. We then apply the pre-trained ResNet50 model on these three images $I, I_{pos}, I_{neg}$ to extract features, denoted as $x, x_{pos}, x_{neg}\in\mathbb{R}^{F}$, respectively. The similarity scores of samples are calculated using a simple bilinear scoring function with sigmoid activation function as follows:
\begin{equation*}
    s_{pos} = \sigma(xWx_{pos}), \quad
    s_{neg} = \sigma(xWx_{neg}),
\end{equation*}
where $W\in \mathbb{R}^{F\times F}$ is learnable parameter, $\sigma$ is the sigmoid function. The pre-trained ResNet50 model is finetuned using the binary cross-entropy loss function. 
Namely, the loss function can be expressed as
\begin{equation}
    L_{cl} = \frac{1}{|I|}\sum_{I}\left( -log(s_{pos}) -log(1-s_{neg})\right).
\end{equation}

We note that the structure of the CL task is similar to Triplet network~\cite{hoffer2015deep}. Intuitively, the task is designed to map images with similar scenes  closely to each other and dissimilar scenes separable as farther apart as possible. Thus images with similar scenes could have high similarity scores and vice versa. This contrastive learning paradigm enables the model to learn discriminative scene representations. 

After finetuning the pre-trained CNN model, we use it as a scene feature extractor in our downstream social relation inference task. Namely, given an image $I$, we obtain the scene feature of the image as follows:
\begin{equation}
    x'_{I} = f_{CL-RoI}(I, b_{I}),
\end{equation}
where $f_{CL-RoI}(\cdot)$ represents the CL finetuned CNN model with RoI pooling layer.

Finally, to predict the relational class distribution of person $i$ and $j$ in the image, we concatenate their interactive feature $r_{ij}$ extracted from RGCN model, foreground feature $x_{ij}$, background feature $x_{I}$ extracted from ImageNet pre-trained CNN model and discriminative scene feature $x'_{I}$ extracted from CL finetuned model together. The concatenated features are fed as input to the MLP layer for relation classification. The network outputs relational class distribution for all person pairs in the image. 

We note that most of the previous methods, such as Pair CNN, Dual-Glance, SRG-GN, etc, consider the social relations on the same image separately. Namely, their model outputs relational class distribution for single pair of person \textit{only}, even if there are multiple people in the image. This may cause occurrence of some obviously unreasonable and contradictory relationships in one image. In contrast, our model directly learns the joint distribution of social relations for multiple people. Given an image as input, PRISE extracts features of multiple people in the image and directly outputs the relational class distributions for \textit{all} person pairs. This enables our model to generate reasonable and consistent social relationships in one image.

\section{Experiments}
In this section, we conduct extensive experiments based on PIPA and PISC, as well as a new large-scale unlabeled dataset. We first present the description of datasets and the implementation details. Then we evaluate the performance of our proposed model through comparisons with benchmarks and ablation study. Finally, we visualize the results from the CL model with discussions. All the codes and experimental results are publicly available on github\footnote{https://github.com/IFBigData/PRISE}.

\subsection{Datasets}
\noindent\textbf{Social Relation Datasets.} We conduct experiments on two social relation datasets, i.e., the PIPA dataset~\cite{zhang2015beyond} and the PISC dataset~\cite{li2017dual}. 
The \textit{accuracy} over all classes is used to evaluate all methods in PIPA dataset.
The PISC dataset has a hierarchy of three coarse-level relations (intimate, non-intimate, no relation) and six fine-level relations (friend, family, couple, professional, commercial, and no relation). 
For fair comparisons, we follow the standard train/val/test split in~\cite{li2017dual}. 
The \textit{per-class recalls} and \textit{mean Average Precision (mAP)} are used to evaluate all methods in PISC dataset. 

\noindent\textbf{Contrastive Learning Dataset.} For CL, we extend PISC dataset to a new dataset with 240,200 images by using google image search engine. Specifically, we search for 10 similar images on Google for every image in PISC dataset, thus extending PISC dataset by approximately 10 times. We combine the extended dataset with PIPA and PISC images and name it as PISC-extension dataset. We show some examples from this dataset in supplementary materials. 
We take 80\% of samples in the PISC-extension dataset as the training set, and the remaining samples as the test set. 

\subsection{Implementation Details}
\noindent \textbf{Contrastive Learning.} We use the pre-trained ResNet50 model~\cite{zhou2017places} to obtain the top-5 scene categories of an input image. We then construct positive and negative sample pairs based on the scene category. 
Each image has a pool of positive samples and a pool of negative samples. For simplicity, we limit the maximum number of images in a pool for each image. In this paper, we set the maximum number as 50~\footnote{We have considered other values (e.g., 30, and 80) and found that this parameter is insensitive to the results.}. 
In the training phase, we randomly select one image from the positive and negative sample pool respectively. We set the batch size to be 32, the learning rate to be $1\times 10^{-5}$. The ResNet50 model is finetuned end-to-end using the Adam optimizer. For the performance of the finetuned ResNet50 model, the accuracy and AUC on the test set are 91.0\% and 96.7\%, respectively. After training, the network parameters are saved for the downstream task.
  
\noindent\textbf{Training of PRISE.} Our PRISE is trained with a learning rate of $5\times 10^{-5}$. We resize the input image into $448\times 448$, and train the network for 20 epochs with a batch size of 32. The number of layers in RGCN is set to be 2, i.e., $T=2$. 

\subsection{Comparisons with Benchmarks}

\begin{table}[t]
    \centering
    \caption{Comparisons of the accuracy (in \%) between our PRISE and other state-of-the-art methods in PIPA dataset.}
    \begin{tabular}{l|c|c}
    \toprule
        Methods & domain & relation \\
        \midrule
        Pair CNN~\cite{li2017dual} & 65.9 & 58.0 \\
        Dual-Glance~\cite{li2017dual} & - & 59.6 \\
        SRG-GN~\cite{goel2019end} & - & 53.6 \\
        GRM~\cite{wang2018deep} & - & 62.3 \\
        MGR~\cite{zhang2019multi} & - & 64.4 \\
        GR$^2$N~\cite{li2020graph} & 72.3 & 64.3\\
        \midrule
        PRISE &  \textbf{77.2} & \textbf{69.5} \\
    \bottomrule
    \end{tabular}
    \label{tab:pipa}
\end{table}

In experiments, we compare PRISE with the following existing methods. For fair comparisons, we report the best results in experiments for Tables~\ref{tab:pipa} and ~\ref{tab:pisc} following the routine in this research field.

  \textbf{Pair CNN}~\cite{li2017dual}. Two cropped image patches of the two persons are fed into two CNNs with sharing weights to extracted features for social relation classification.
  
  \textbf{Dual-Glance}~\cite{li2017dual}. The first glance focuses on the pair of people. The second glance extracts the information of objects in the context to refine the prediction.
  
  \textbf{SRG-GN}~\cite{goel2019end}. Scene and human attribute context features are extracted by five CNNs. 
  
  \textbf{GRM}~\cite{wang2018deep}. This model represents the person and objects existing in an image as a weighted graph, and then using a gated graph network to predict social relation.
  
  \textbf{MGR}~\cite{zhang2019multi}. This model employs two graph neural network (GNN) to extract the relationship between people and the relationship between people and objects.
  
  \textbf{GR$^2$N}~\cite{li2020graph}. This model uses GNN to model all relationships in one graph which can provide strong logical constraints among different types of social relations.

It is worth noting that all of the above methods, except GR$^2$N, are person pair-based, which means that they consider the pair-wised social relations on the same image separately. In contrast, both PRISE and GR$^2$N consider the social relations among all people in one image jointly. Unlike our method, GR$^2$N do not use foreground, background and scene features. 
Besides, Dual-Glance, GRM and MGR use object information in an image to assist in relation inference. We note that in SRG-GN, they directly apply the pre-trained scene classification model as feature extractor for foreground ground information, while in our PRISE, we first use the CL approach to finetune the pre-trained model, and then apply the model for scene feature extraction. 

The experimental results of social domain recognition and social relationship recognition in PIPA dataset are shown in Table~\ref{tab:pipa}.
We observe that our PRISE outperforms other methods by a significant margin. Specifically, our method achieves an accuracy of $77.2\%$ for social domain recognition and $69.5\%$ for social relation recognition, beating all the person pair-based methods. This shows the benefit of graph-based approach that jointly models all the social relationships among people in an image.
Besides, our method improves the current state-of-the-art method, i.e., GR$^2$N, by $6.8\%$ for social domain recognition and $8.1\%$ for social relation recognition, respectively. 

Similar results can be found in Table~\ref{tab:pisc}, where we shows the experimental comparison with prior methods in PISC dataset. We observe that our method achieves an mAP of $83.4\%$  for the coarse-level recognition and $73.8\%$ for the fine-level recognition, which are new state-of-the-art. We note that PRISE
takes full advantage of holistic scene and thus makes better predictions for non-intimate relation. For degradation in `Int' and `Fri', we argue that the similar scenes of `Fri' and `Fam' misleads our model.

Compared with GR$^2$N, PRISE achieves competitive performance for both coarse and fine relationship recognition with a much simpler GCN structure.
The above results highlight the benefits of concise interaction graph and discriminative scene feature in PRISE. 

\begin{table*}[t]
    \centering
    \caption{Comparisons of the per-class recall for each relationship and the mAP over all relationship (in \%) between our PRISE and other state-of-the-art methods in PISC dataset. Int: Intimate, Non: Non-Intimate, NoR: No Relation, Fri: Friend, Fam: Family, Cou: Couple, Pro: Professional, Com: Commercial, NoR: No Relation.}
    \begin{tabular}{l|c c c|c|c c c c c c|c}
    \toprule
        \multirow{2}{*}{Methods} & \multicolumn{4}{c|}{Coarse relationships} & \multicolumn{7}{c}{Fine relationships}  \\
        \cline{2-12}
        & Int & Non & NoR & mAP & Fri & Fam & Cou & Pro & Com & NoR & mAP \\
        \midrule
        Pair CNN~\cite{li2017dual} & 70.3 & 80.5 & 38.8 & 65.1 & 30.2 & 59.1 & 69.4 & 57.5 & 41.9 & 34.2 & 48.2\\
        Dual-Glance~\cite{li2017dual} & 73.1 & 84.2 & 59.6 & 79.7 & 34.4 & 68.1 & 76.3 & 70.3 & 57.6 & 60.9 & 63.2\\
        SRG-GN~\cite{goel2019end}  & - & - & - & - & - & - & - & - & - & - & 71.6 \\
        GRM~\cite{wang2018deep} & 81.7 & 73.4 & 65.5 & 82.8 & 59.6 & 64.4 & 58.6 & 76.6 & 39.5 & 67.7 & 68.7\\
        MGR~\cite{zhang2019multi} & - & - & - & - & 64.6 & 67.8 & 60.5 & 76.8 & 34.7 & 70.4 & 70.0\\
        GR$^2$N~\cite{li2020graph}& 81.6 & 74.3 & 70.8 & 83.1 & 60.8 & 65.9 & 84.8 & 73.0 & 51.7 & 70.4 & 72.7\\
        \midrule
        PRISE & 73.3 & 79.2 & 71.8 & \textbf{83.4} & 47.1 & 74.7 & 76.6 & 73.2 & 70.3 & 68.2 & \textbf{73.8} \\
    \bottomrule
    \end{tabular}
    \label{tab:pisc}
\end{table*}


\subsection{Ablation Study}
We conduct ablation study to show how much each component of PRISE contributes to the performance.
Specifically, we remove the interactive feature, the scene feature, the foreground and background features from PRISE, denoted as \textit{w/o Int.}, \textit{w/o Scene}, \textit{w/o Fore.}, \textit{w/o Back.}, respectively. 
In addition, to show the effectiveness of discriminative scene representation, we consider a variant denoted by \textit{PRISE$|$Pretrained}, where we replace the CL finetuned model with the ResNet50 that was pretrained on Place365 dataset.
The results are summarized in Table~\ref{tab:ablation_study}. 

As we can see in Table~\ref{tab:ablation_study}, among all the four components, the interactive feature is the most important, followed by the scene feature. Without interactive feature, the mean of mAP in PISC-coarse and PISC-Fine dataset drops 7.4\% and 11.3\% in absolute value, respectively. These two numbers become 0.8\% and 0.9\% if we remove the scene feature from PRISE.
On one hand, this result demonstrates the effectiveness of RGCN to extract interactive feature. On the other hand, it shows the benefit of considering scene information in social relation understanding.

In order to give the full picture of the efficacy of the proposed method, we slowly add single type of features to conduct ablation study. We present the results in Table~\ref{tab:ablation_study_v2}. We can observe that all of the four features contribute to the inference of social relation.

Besides, by comparing the results of \textit{PRISE$|$Pretrained} and PRISE in Table~\ref{tab:ablation_study}, we clearly find that the discriminative scene representation provides significant hints for social relation classifications, especially for fine relationships with $1.8\%$ relative improvement.

The PISC-extension dataset is used in CL to finetune the scene feature extractor. With this newly added dataset, we can obtain scene representations that are more discriminative and thus useful for downstream task. For instance, in PISC-Fine dataset, using PISC-extension dataset can boost the mAP from $72.0\pm0.4$ to $72.8\pm0.5$.

\begin{table}[t]
    \centering
    \caption{Ablation study of the PRISE model in PISC dataset. We report the mean and standard deviation of mAP (in \%) among 50 random runs in PISC dataset.}
    \begin{tabular}{l|c|c}
    \toprule
        Methods & Coarse & Fine\\
        \midrule 
        PRISE w/o Int. &  $75.3\pm0.2$ & $61.4\pm0.4$\\
        PRISE w/o Scene & $81.9\pm0.3$ & $71.8\pm0.4$\\
        PRISE w/o Fore. &  $82.2\pm0.4$ & $71.9\pm0.5$\\
        PRISE w/o Back.  & $82.5\pm0.3$ & $72.5\pm0.4$\\
        \midrule
        PRISE$|$Pretrained   & $82.2\pm0.4$ & $71.4\pm0.4$\\
        \midrule
        PRISE &  \boldmath{$82.8\pm0.3$} & \boldmath{$72.8\pm0.5$}\\
    \bottomrule
    \end{tabular}
    \label{tab:ablation_study}
\end{table}

\begin{table}[t]
    \centering
    \caption{Ablation study of PRISE. We report the mean and standard deviation of mAP (in \%) among 50 random runs in PISC.}
    \begin{tabular}{l|c|c}
    \hline
        Model Structure &  Coarse & Fine \\
       \hline
        Int. &  $80.8 \pm0.3$ & $70.9\pm0.4$\\
        \hline
        Int.+Scene &  $81.6 \pm0.4$& $71.6\pm0.4$\\
        \hline
        Int.+Scene+Fore. & $82.5\pm0.3$ &   $72.5\pm0.4$\\
        \hline
       Int.+Scene+Fore.+Back.  &  $\textbf{82.8} \pm 0.3$ & $\textbf{72.8} \pm 0.5$\\
    \hline
    \end{tabular}
    \label{tab:ablation_study_v2}
\end{table}

\subsection{Visualization of Scene Representations}

\begin{figure}[ht]
 \begin{center}
 \includegraphics[width=0.85\columnwidth]{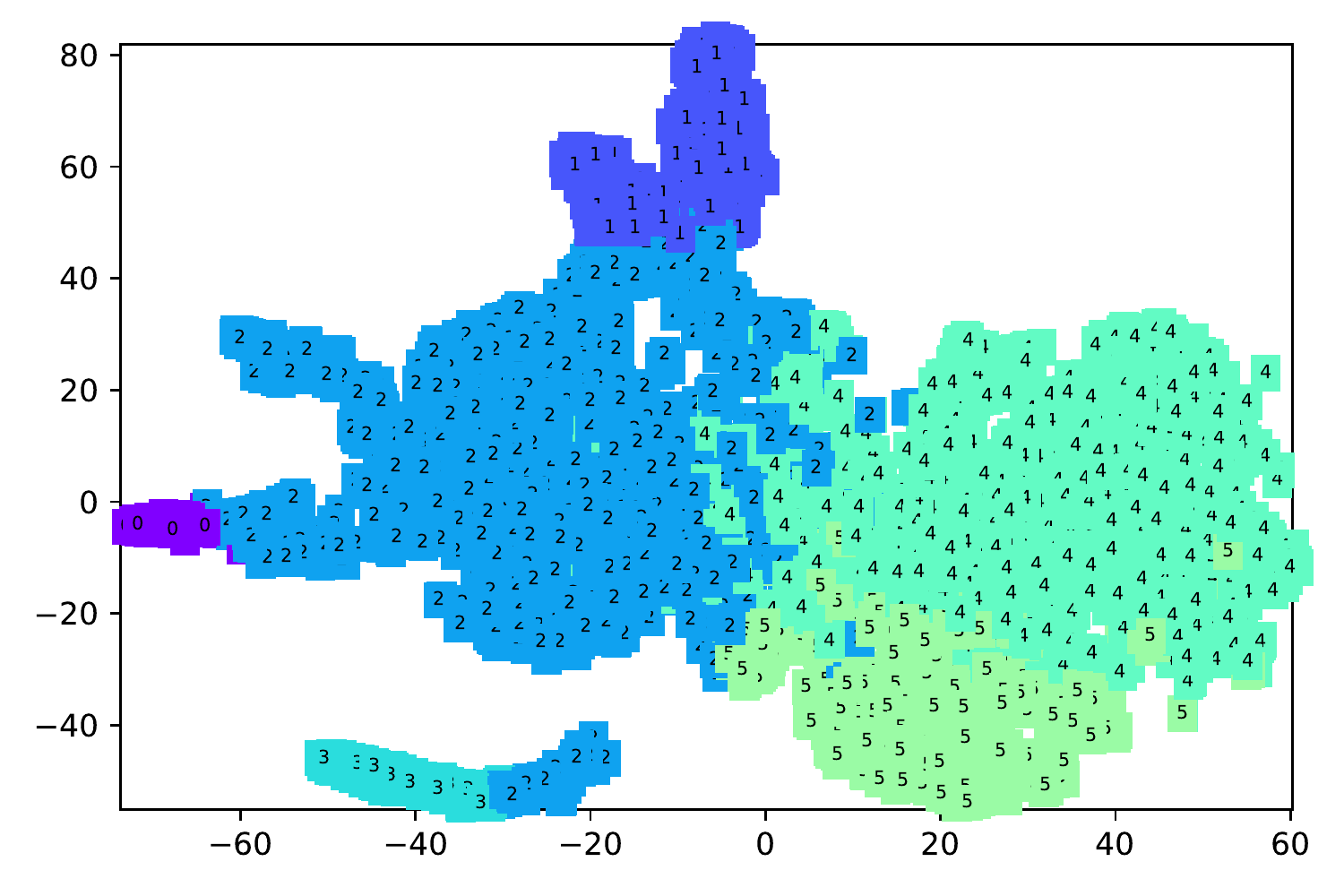}
 \end{center}
 \caption{\label{fig:ssl_cluster_tsne_new} TSNE visualization of image scene features obtained from CL finetuned model on PISC test dataset. Different colors are used to represent clusters of different categories.}
\end{figure}

To demonstrate the discriminative scene representation learned by our CL finetuned model, we randomly choose 4000 images from PISC-coarse test dataset, and conduct a clustering task based on the learned features. Specifically, we first use the CL finetuned model to generate the 2048-dimensional scene representations for each image. Then, we use spectral clustering to cluster these features. The 4000 test images are divided into 6 categories. We visualize sample images in different clusters in supplementary materials. We use TSNE to reduce the features dimension from 2048 to 2, and visualize them in Figure~\ref{fig:ssl_cluster_tsne_new}. We can observe that images from different categories are separated. These results directly show that the features learned by our 
CL finetuned model are discriminative. 

\subsection{Discussions on Effectiveness of PRISE}

\noindent\textbf{RGCN as Interactive Feature Extractor: Simpler and Faster.}
We would have used GR$^2$N~\cite{li2020graph} as the interactive feature extractor, however, we note that it is too complicated. 
For a social relation inference problem with $K$ categories of social relation, GR$^2$N introduced $K$ sets of trainable parameters. In contrast, the number of parameters of our RGCN does not depend on the number of social relation categories, which makes RGCN much simpler. 

To further compare GR$^2$N and RGCN as interactive feature extractors in terms of performance and inference time, we conduct experiments by replacing RGCN with GR$^2$N in PRISE while fixing other components. Specifically, in GR$^2$N each category of social relation has a representation. We apply a max pooling operator on representations of different social relations to obtain the interactive feature, and replace $r_{ij}$ in our PRISE with this new interactive feature. We denote this setting as \textit{PRISE$|$GR$^2$N}. The experimental results on the accuracy and inference time~\footnote{The inference time reported here does not include the time needed to extract features using ResNet101 model and CL finetuned model.} on test set of both algorithms in PIPA dataset are shown in Table~\ref{tab:pipa_gr2n_rgcn}.  We can observe that the performance of PRISE$|$GR$^2$N is comparable to PRISE. However, in terms of inference time, the model with GR$^2$N is much slower as compared to PRISE. These results strongly support our conclusion that the proposed RGCN model as interactive feature extractor is much simpler and faster as compared to GR$^2$N.

\begin{table}[t]
    \centering
    \caption{Comparisons of GR$^2$N and RGCN in PIPA dataset. We report the accuracy (in \%) and inference time.}
    \begin{tabular}{l|c c|c c}
    \toprule
        \multirow{2}{*}{Methods} & 
        \multicolumn{2}{c|}{domain} &
        \multicolumn{2}{c}{relation} \\
        \cline{2-5}
        & accuracy & time & accuracy & time \\
        \midrule
        PRISE$|$GR$^2$N & 75.6 & 3.33s & 68.1 & 6.23s\\
        \midrule
        PRISE &  \textbf{77.2} & 1.91s & \textbf{69.5} & 1.82s\\
    \bottomrule
    \end{tabular}
    \label{tab:pipa_gr2n_rgcn}
\end{table}

\begin{figure}[ht]
 \begin{center}
 \includegraphics[width=0.8\columnwidth]{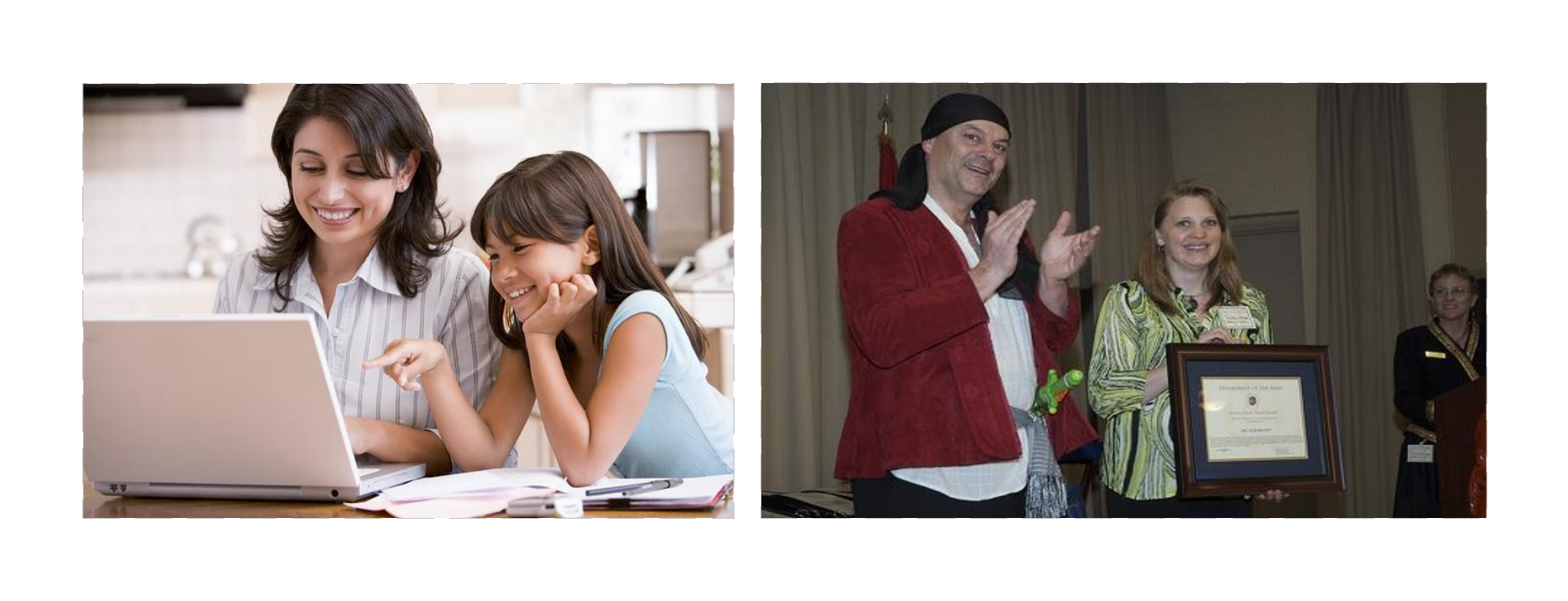}
 \end{center}
 \caption{\label{fig:sample_image} Visualization of sample images from PISC test dataset, where \textit{PRISE w/o Scene} makes wrong predictions while PRISE makes correct predictions. }
\end{figure}

\noindent \textbf{Benefits of Utilizing the Scene Information.} Scene information is important for social relation understanding. In Figure~\ref{fig:sample_image}, we visualize two sample images from PISC test dataset, where \textit{PRISE w/o Scene} makes wrong predictions while PRISE makes correct predictions. In this example, people (in the left image) in a home environment tends to have intimate relation, while people (in the right image) in public environment tends to have non-intimate relation. Without scene information, PRISE makes wrong predictions by predicting the two persons in the left image to have non-intimate relation, and the two persons in the right image to have intimate relation. More examples are illustrated in supplementary materials. 

\begin{figure}[ht]
 \begin{center}
 \includegraphics[width=0.8\columnwidth]{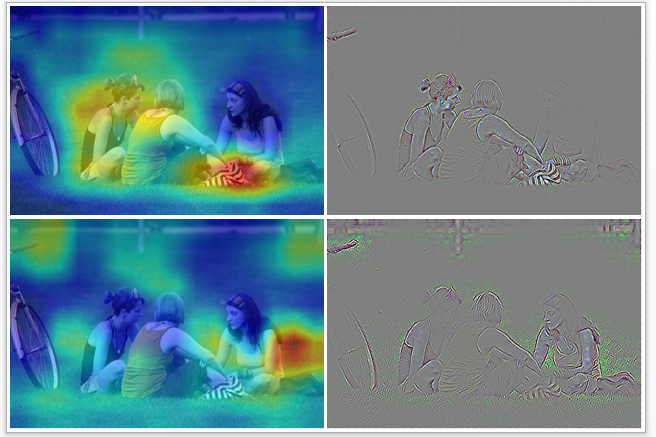}
 \end{center}
 \caption{\label{fig:heatmap} Heat map and gradient map  of scene representations extracted from different models. The top two images are the pre-trained model on Imagenet. The bottom two images are from the pre-trained model on Place365. }
\end{figure}

Besides, the scene representations extracted from different models could be very different. In Figure~\ref{fig:heatmap}, we show the heat map and gradient map~\cite{selvaraju2017grad} of the pre-trained models from ImageNet and Place365~\cite{zhou2017places} dataset, respectively. We can observe that the feature extracted by ImageNet pre-trained model focuses more on the persons, while the feature extracted by Place365 pre-trained model focuses more on the holistic scene. 

\section{Conclusion}
In this paper, we have originally proposed  PRISE to enhance social relation inference. PRISE synthesizes three streams of information, i.e., holistic scenes, foreground and background information of persons and objects, and interaction of persons. Technically, we have developed a RGCN model to extract interactive features and designed a CL task to learn discriminative scene representations. The RGCN model in PRISE is concise in terms of learning the interaction for all persons in an image and the running time of feature extraction. 
Extensive experiments demonstrate that PRISE is superior than prior methods, and achieves new state-of-the-art results in PIPA and PISC datasets. The contrastive learning task in PRISE sheds new lights on improving performance of more complicated tasks in computer vision, such as behaviour analysis and image captioning.

\appendix
\section{Supplementary Material}
In this supplementary material, we show experimental results to further demonstrate the effectiveness and superiority of our proposed PRISE model. We first show the visualization of PISC-extension dataset in Section~\ref{sec:visualization_of_pisc-extension}. In Section~\ref{sec:visualization_of_ssl}, we visualize the scene representations learned by the contrastive learning finetuned model. We show the sample images from PISC dataset to illustrate the benefits of holistic scene feature in Section~\ref{sec:benefit_of_ssl}. Finally, we conduct extensive ablation study on PIPA dataset to illustrate the benefits of each components of PRISE in Section~\ref{sec:ablation_study_pipa}.  To reproduce all the experimental results in this work, \textbf{our proposed PRISE and the parameter settings are available on github\footnote{https://github.com/IFBigData/PRISE}}.

\section{Visualization of PISC-extension dataset}\label{sec:visualization_of_pisc-extension}
\begin{figure}[h]
 \begin{center}
 \includegraphics[width=0.95\columnwidth]{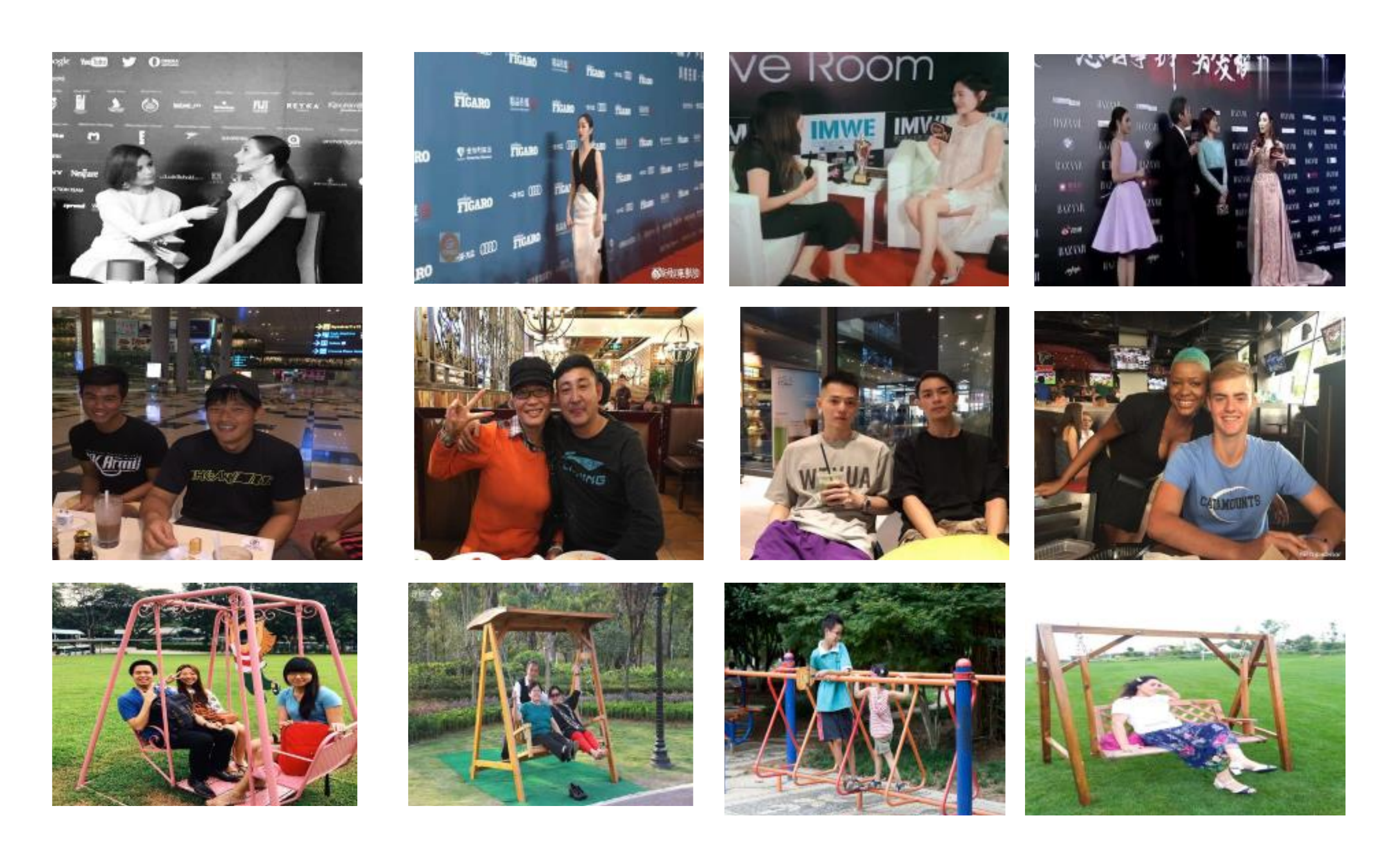}
 \end{center}
 \caption{\label{fig:pisc-extension} Visualization of the PISC-extension dataset. The images in the first column are from PISC dataset, and images from following columns are from Google image search engine.}
\end{figure}

For contrastive learning, we extend PISC dataset to a new dataset with 240,200 images by using google image search engine. Specifically, we search for 10 similar images on Google for every image in PISC dataset, thus extending PISC dataset by approximately 10 times. We combine the extended dataset with PIPA and PISC images and name it as PISC-extension dataset. We show some examples from this dataset in Figure~\ref{fig:pisc-extension}. As can be seen, the images from google image search engine are similar to the original images from PISC dataset.

\section{Visualization of scene representation}\label{sec:visualization_of_ssl}

\begin{figure}[ht]
 \begin{center}
 \includegraphics[width=0.95\columnwidth]{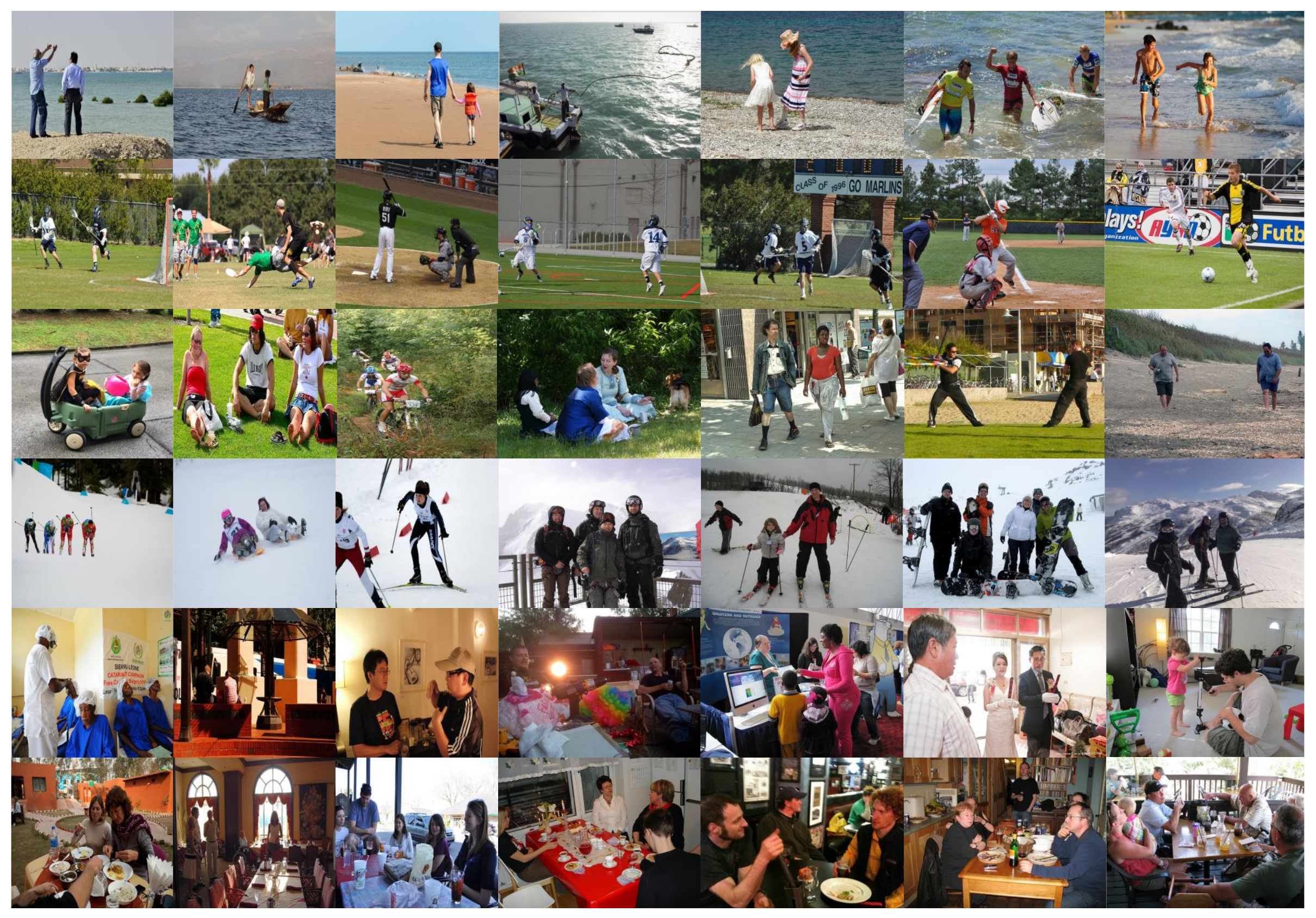}
 \end{center}
 \caption{\label{fig:ssl_cluster_image} Visualization of the images in different clusters. Each row of images belongs to the same cluster and the $i$-th (i=0,1,2,3,4,5) row corresponds to the $i$-th category in Figure~\ref{fig:ssl_cluster_tsne_new}.}
\end{figure}

To further demonstrate the scene information learned by our contrastive learning finetuned model, we randomly choose 4000 images from PISC test dataset in the coarse-relation domain, and conduct a clustering task based on the learned features. Specifically, we first use the finetuned model to generate the 2048-dimensional features for each image. Then, we use spectral clustering to cluster these features. The 4000 test images are divided into 6 categories. We use TSNE to reduce the features dimension from 2048 to 2, and visualize them in Figure~\ref{fig:ssl_cluster_tsne_new}. We can observe that images from different categories are separated.
We randomly select 7 images from each category and display them in Figure~\ref{fig:ssl_cluster_image}. Overall, we can observe that the similarity between pictures in each category is relatively high, while the similarity between different categories is relatively low. 

In Figure~\ref{fig:ssl_cluster_image}, each row of images belongs to the same cluster and the $i$-th (i=0,1,2,3,4,5) row corresponds to the $i$-th category in Figure~\ref{fig:ssl_cluster_tsne_new}. For more details, category 0 contains a total of 52 images, all of which are scenes of playing at the beach; category 1 contains a total of 348 images, all of which are sport games on the grass; category 2 contains a total of 1565 images, most of which are some outdoor leisure activities; category 3 contains a total of 118 images, all of which are scenes of activities on the snow; category 4 contains a total of 1400 images, most of which are indoor scenes; category 5 contains 517 images, most of which are related to eating scene.

Based on these results, we conclude that the features extracted by our contrastive learning finetuned model contain rich and discriminative scene information.

\section{Benefits of contrastive learning finetuned model: fully utilizing the holistic scene feature}\label{sec:benefit_of_ssl}

\begin{figure}[ht]
 \begin{center}
 \includegraphics[width=0.8\columnwidth]{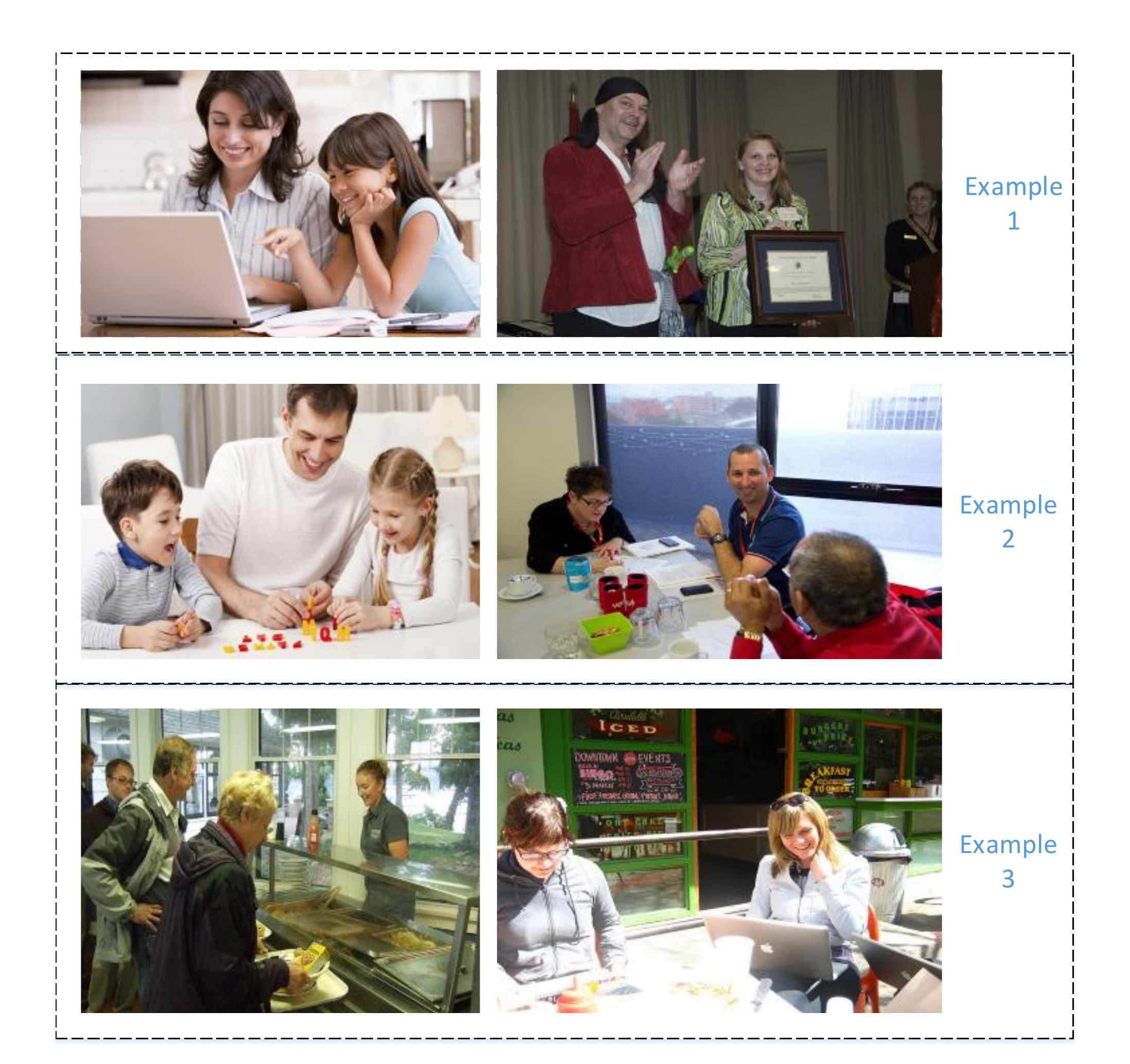}
 \end{center}
 \caption{\label{fig:ssl_result} Visualization of sample images from PISC test dataset, where PRISE without scene information makes wrong predictions while PRISE makes correct predictions. }
\end{figure}

In Figure~\ref{fig:ssl_result}, we visualize sample images from PISC test dataset, where PRISE without scene information makes wrong predictions while PRISE makes correct predictions. All of these sample images show the importance of holistic scene information for social relation inference. Specifically, in the first example in Figure~\ref{fig:ssl_result}, people (in the left image) in a home environment tends to have intimate relation, while people (in the right image) in public environment tends to have non-intimate relation. Without scene information, the PRISE makes wrong predictions by predicting the two persons in the left image to have non-intimate relation, and the two persons in the right image to have intimate relation. 
Similarly in the second example in Figure~\ref{fig:ssl_result}, people (in the left image) in a home environment tends to have intimate relation, while people (in the right image) in an office environment tends to have non-intimate relation. Without scene information, the PRISE makes wrong predictions by predicting all the people in the left image to have no relation, and all the people in the right image to have intimate relation. In the third example in Figure~\ref{fig:ssl_result}, the seller and the customers in the canteen (in the left image) have non-intimate relation, while people (in the right image) having lunch on the same table are intimate relation. Without scene information, the PRISE makes wrong inference by predicting the relation between seller and customer in the left image to have intimate relation, and people in the right image to have non-intimate relation.

\section{Ablation study on PIPA}\label{sec:ablation_study_pipa}
Similar to the main body, we conduct ablation study to show how much each component of PRISE contributes to the performance.
Specifically, we remove the interactive feature, the scene feature, the foreground and background features from PRISE, denoted as \textit{w/o Int.}, \textit{w/o Scene}, \textit{w/o Fore.}, \textit{w/o Back.}, respectively. 
In addition, to show the effectiveness of discriminative scene representation, we consider a variant denoted by \textit{PRISE$|$Pretrained}, where we replace the contrastive learning finetuned model with the ResNet50 that was pretrained on Place365 dataset.
The results on PIPA dataset are summarized in Table~\ref{tab:ablation_study_pipa}. 

\begin{figure}[ht]
 \begin{center}
 \includegraphics[width=0.8\columnwidth]{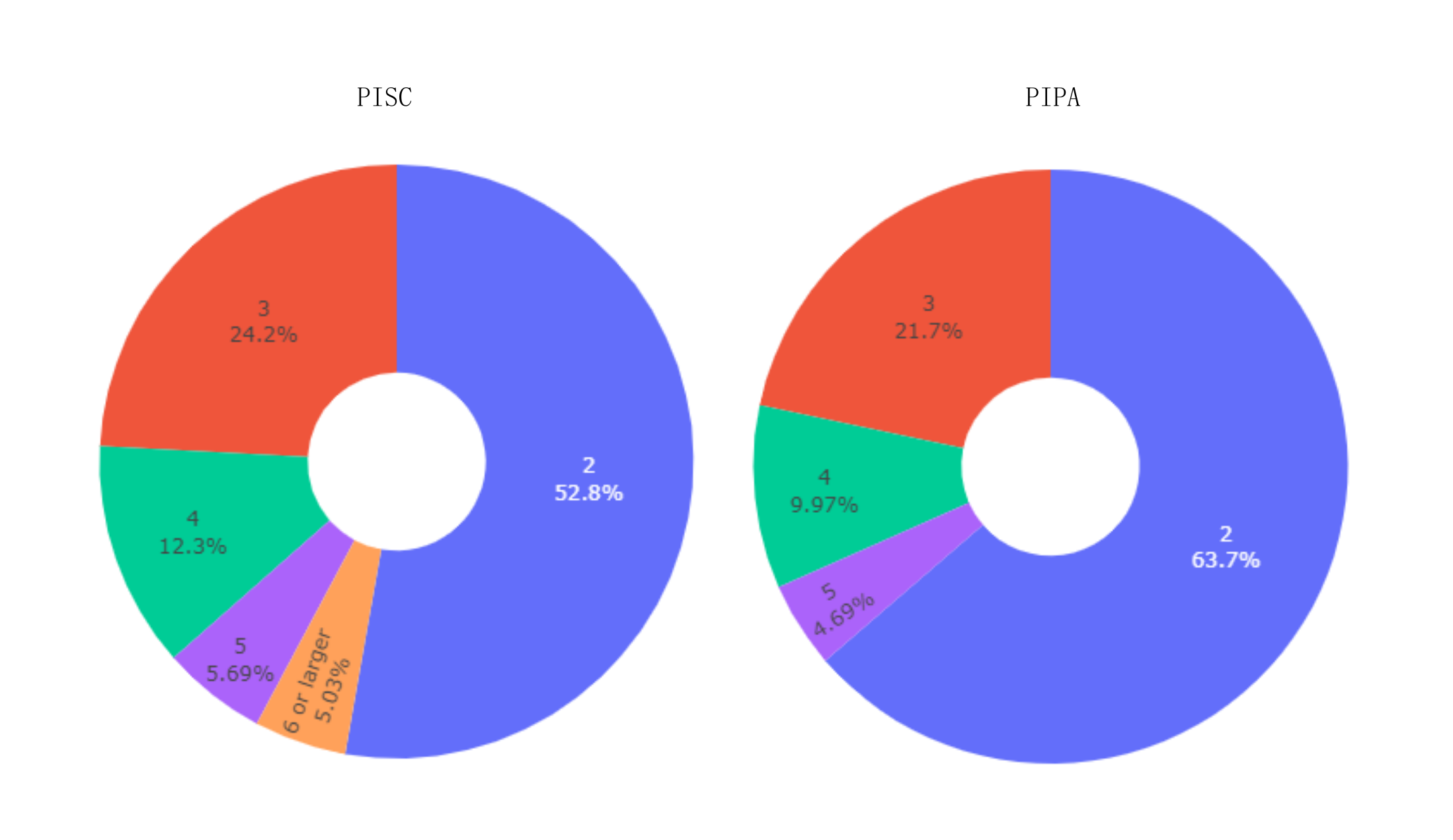}
 \end{center}
 \caption{\label{fig:person_count} Pie chart of number of person per image in PISC and PIPA dataset. The average numbers of persons per image in PISC and PIPA dataset are 2.89 and 2.56, respectively.}
\end{figure}

As we can see in Table~\ref{tab:ablation_study_pipa}, among all the four components, the interactive feature and the scene feature are the most important. Without interactive feature, the mean of accuracy in PIPA-domain and PIPA-relation dataset drops 2.2\% and 0.3\% in absolute value, respectively. These two numbers become 2.5\% and 0.8\% if we remove the scene feature from PRISE. We note that in PIPA dataset, the average number of persons in each image is lower than that in PISC dataset (as shown in Figure~\ref{fig:person_count}), probably leading to the mild difference in the ablation study.
\begin{table}[t]
    \centering
    \caption{Ablation study of the PRISE model in PIPA dataset. We report the mean and standard deviation of accuracy (in \%) among 50 random runs in PIPA dataset.}
    \begin{tabular}{l|c|c}
    \toprule
        Methods & Domain & Relation\\
        \midrule 
        PRISE w/o Int. &  $72.9 \pm0.4$ & $66.5\pm0.5$\\
        PRISE w/o Scene & $72.6\pm0.3$ & $66.0\pm0.7$\\
        PRISE w/o Fore. &  $73.2\pm0.4$ & $66.7\pm0.7$\\
        PRISE w/o Back.  & $73.2\pm0.4$ & $66.9\pm0.6$\\
        \midrule
        PRISE$|$Pretrained   & $73.3\pm0.2$ & $66.0\pm0.6$\\
        \midrule
        PRISE &  $75.1\pm0.6$ & $66.8\pm0.8$\\
    \bottomrule
    \end{tabular}
    \label{tab:ablation_study_pipa}
\end{table}

\bibliographystyle{unsrt}  
\bibliography{references}  

\begin{thebibliography}{10}

\bibitem{kitayama2000pursuit}
Shinobu Kitayama and Hazel~Rose Markus.
\newblock The pursuit of happiness and the realization of sympathy: Cultural
  patterns of self, social relations, and well-being.
\newblock {\em Culture and subjective well-being}, 1:113--161, 2000.

\bibitem{bromley1988understanding}
David~G Bromley and Bruce~C Busching.
\newblock Understanding the structure of contractual and covenantal social
  relations: Implications for the sociology of religion.
\newblock {\em SA. Sociological analysis}, pages 15S--32S, 1988.

\bibitem{hoai2014talking}
Minh Hoai and Andrew Zisserman.
\newblock Talking heads: Detecting humans and recognizing their interactions.
\newblock In {\em Proceedings of the IEEE Conference on Computer Vision and
  Pattern Recognition}, pages 875--882, 2014.

\bibitem{khademi2018image}
Mahmoud Khademi and Oliver Schulte.
\newblock Image caption generation with hierarchical contextual visual spatial
  attention.
\newblock In {\em Proceedings of the IEEE Conference on Computer Vision and
  Pattern Recognition Workshops}, pages 1943--1951, 2018.

\bibitem{alahi2016social}
Alexandre Alahi, Kratarth Goel, Vignesh Ramanathan, Alexandre Robicquet,
  Li~Fei-Fei, and Silvio Savarese.
\newblock Social lstm: Human trajectory prediction in crowded spaces.
\newblock In {\em Proceedings of the IEEE conference on computer vision and
  pattern recognition}, pages 961--971, 2016.

\bibitem{li2017dual}
Junnan Li, Yongkang Wong, Qi~Zhao, and Mohan~S Kankanhalli.
\newblock Dual-glance model for deciphering social relationships.
\newblock In {\em Proceedings of the IEEE International Conference on Computer
  Vision}, pages 2650--2659, 2017.

\bibitem{goel2019end}
Arushi Goel, Keng~Teck Ma, and Cheston Tan.
\newblock An end-to-end network for generating social relationship graphs.
\newblock In {\em Proceedings of the IEEE Conference on Computer Vision and
  Pattern Recognition}, pages 11186--11195, 2019.

\bibitem{zhang2019multi}
Meng Zhang, Xinchen Liu, Wu~Liu, Anfu Zhou, Huadong Ma, and Tao Mei.
\newblock Multi-granularity reasoning for social relation recognition from
  images.
\newblock In {\em 2019 IEEE International Conference on Multimedia and Expo
  (ICME)}, pages 1618--1623. IEEE, 2019.

\bibitem{li2020graph}
Wanhua Li, Yueqi Duan, Jiwen Lu, Jianjiang Feng, and Jie Zhou.
\newblock Graph-based social relation reasoning.
\newblock {\em arXiv preprint arXiv:2007.07453}, 2020.

\bibitem{wang2018deep}
Zhouxia Wang, Tianshui Chen, Jimmy Ren, Weihao Yu, Hui Cheng, and Liang Lin.
\newblock Deep reasoning with knowledge graph for social relationship
  understanding.
\newblock {\em arXiv preprint arXiv:1807.00504}, 2018.

\bibitem{shu2019hierarchical}
Xiangbo Shu, Jinhui Tang, Guojun Qi, Wei Liu, and Jian Yang.
\newblock Hierarchical long short-term concurrent memory for human interaction
  recognition.
\newblock {\em IEEE transactions on pattern analysis and machine intelligence},
  2019.

\bibitem{lu2017discriminative}
Jiwen Lu, Junlin Hu, and Yap-Peng Tan.
\newblock Discriminative deep metric learning for face and kinship
  verification.
\newblock {\em IEEE Transactions on Image Processing}, 26(9):4269--4282, 2017.

\bibitem{liang2018weighted}
Jianqing Liang, Qinghua Hu, Chuangyin Dang, and Wangmeng Zuo.
\newblock Weighted graph embedding-based metric learning for kinship
  verification.
\newblock {\em IEEE Transactions on Image Processing}, 28(3):1149--1162, 2018.

\bibitem{lu2013neighborhood}
Jiwen Lu, Xiuzhuang Zhou, Yap-Pen Tan, Yuanyuan Shang, and Jie Zhou.
\newblock Neighborhood repulsed metric learning for kinship verification.
\newblock {\em IEEE transactions on pattern analysis and machine intelligence},
  36(2):331--345, 2013.

\bibitem{xu2015show}
Kelvin Xu, Jimmy Ba, Ryan Kiros, Kyunghyun Cho, Aaron Courville, Ruslan
  Salakhudinov, Rich Zemel, and Yoshua Bengio.
\newblock Show, attend and tell: Neural image caption generation with visual
  attention.
\newblock In {\em International conference on machine learning}, pages
  2048--2057, 2015.

\bibitem{wang2010seeing}
Gang Wang, Andrew Gallagher, Jiebo Luo, and David Forsyth.
\newblock Seeing people in social context: Recognizing people and social
  relationships.
\newblock In {\em European conference on computer vision}, pages 169--182.
  Springer, 2010.

\bibitem{zhang2015learning}
Zhanpeng Zhang, Ping Luo, Chen-Change Loy, and Xiaoou Tang.
\newblock Learning social relation traits from face images.
\newblock In {\em Proceedings of the IEEE International Conference on Computer
  Vision}, pages 3631--3639, 2015.

\bibitem{hess2000influence}
Ursula Hess, Sylvie Blairy, and Robert~E Kleck.
\newblock The influence of facial emotion displays, gender, and ethnicity on
  judgments of dominance and affiliation.
\newblock {\em Journal of Nonverbal behavior}, 24(4):265--283, 2000.

\bibitem{gottman2001facial}
John Gottman, Robert Levenson, and Erica Woodin.
\newblock Facial expressions during marital conflict.
\newblock {\em Journal of Family Communication}, 1(1):37--57, 2001.

\bibitem{zhang2015beyond}
Ning Zhang, Manohar Paluri, Yaniv Taigman, Rob Fergus, and Lubomir Bourdev.
\newblock Beyond frontal faces: Improving person recognition using multiple
  cues.
\newblock In {\em Proceedings of the IEEE conference on computer vision and
  pattern recognition}, pages 4804--4813, 2015.

\bibitem{sun2017domain}
Qianru Sun, Bernt Schiele, and Mario Fritz.
\newblock A domain based approach to social relation recognition.
\newblock In {\em Proceedings of the IEEE Conference on Computer Vision and
  Pattern Recognition}, pages 3481--3490, 2017.

\bibitem{kipf2016semi}
Thomas~N. Kipf and Max Welling.
\newblock Semi-supervised classification with graph convolutional networks.
\newblock In {\em ICLR}, 2017.

\bibitem{chen2020simple}
Ting Chen, Simon Kornblith, Mohammad Norouzi, and Geoffrey Hinton.
\newblock A simple framework for contrastive learning of visual
  representations.
\newblock In {\em International conference on machine learning}, pages
  1597--1607. PMLR, 2020.

\bibitem{devlin2018bert}
Jacob Devlin, Ming-Wei Chang, Kenton Lee, and Kristina Toutanova.
\newblock Bert: Pre-training of deep bidirectional transformers for language
  understanding.
\newblock {\em arXiv preprint arXiv:1810.04805}, 2018.

\bibitem{doersch2015unsupervised}
Carl Doersch, Abhinav Gupta, and Alexei~A Efros.
\newblock Unsupervised visual representation learning by context prediction.
\newblock In {\em Proceedings of the IEEE international conference on computer
  vision}, pages 1422--1430, 2015.

\bibitem{pathak2016context}
Deepak Pathak, Philipp Krahenbuhl, Jeff Donahue, Trevor Darrell, and Alexei~A
  Efros.
\newblock Context encoders: Feature learning by inpainting.
\newblock In {\em Proceedings of the IEEE conference on computer vision and
  pattern recognition}, pages 2536--2544, 2016.

\bibitem{xiao2020should}
Tete Xiao, Xiaolong Wang, Alexei~A Efros, and Trevor Darrell.
\newblock What should not be contrastive in contrastive learning.
\newblock {\em arXiv preprint arXiv:2008.05659}, 2020.

\bibitem{yang2019xlnet}
Zhilin Yang, Zihang Dai, Yiming Yang, Jaime Carbonell, Russ~R Salakhutdinov,
  and Quoc~V Le.
\newblock Xlnet: Generalized autoregressive pretraining for language
  understanding.
\newblock {\em Advances in Neural Information Processing Systems},
  32:5753--5763, 2019.

\bibitem{zhang2017split}
Richard Zhang, Phillip Isola, and Alexei~A Efros.
\newblock Split-brain autoencoders: Unsupervised learning by cross-channel
  prediction.
\newblock In {\em Proceedings of the IEEE Conference on Computer Vision and
  Pattern Recognition}, pages 1058--1067, 2017.

\bibitem{chuang2020debiased}
Ching-Yao Chuang, Joshua Robinson, Lin Yen-Chen, Antonio Torralba, and Stefanie
  Jegelka.
\newblock Debiased contrastive learning.
\newblock {\em arXiv preprint arXiv:2007.00224}, 2020.

\bibitem{feng2019self}
Zeyu Feng, Chang Xu, and Dacheng Tao.
\newblock Self-supervised representation learning from multi-domain data.
\newblock In {\em Proceedings of the IEEE/CVF International Conference on
  Computer Vision}, pages 3245--3255, 2019.

\bibitem{lee2019multi}
Wonhee Lee, Joonil Na, and Gunhee Kim.
\newblock Multi-task self-supervised object detection via recycling of bounding
  box annotations.
\newblock In {\em Proceedings of the IEEE/CVF Conference on Computer Vision and
  Pattern Recognition}, pages 4984--4993, 2019.

\bibitem{dwivedi12benchmarking}
Vijay~Prakash Dwivedi, Chaitanya~K Joshi, Thomas Laurent, Yoshua Bengio, and
  Xavier Bresson.
\newblock Benchmarking graph neural networks.
\newblock {\em TSP}, 12:50--500.

\bibitem{bresson2017residual}
Xavier Bresson and Thomas Laurent.
\newblock Residual gated graph convnets.
\newblock {\em ICLR}, 2018.

\bibitem{xu2018representation}
Keyulu Xu, Chengtao Li, Yonglong Tian, Tomohiro Sonobe, Ken-ichi Kawarabayashi,
  and Stefanie Jegelka.
\newblock Representation learning on graphs with jumping knowledge networks.
\newblock {\em arXiv preprint arXiv:1806.03536}, 2018.

\bibitem{zhou2017places}
Bolei Zhou, Agata Lapedriza, Aditya Khosla, Aude Oliva, and Antonio Torralba.
\newblock Places: A 10 million image database for scene recognition.
\newblock {\em IEEE Transactions on Pattern Analysis and Machine Intelligence},
  2017.

\bibitem{hoffer2015deep}
Elad Hoffer and Nir Ailon.
\newblock Deep metric learning using triplet network.
\newblock In {\em International workshop on similarity-based pattern
  recognition}, pages 84--92. Springer, 2015.

\bibitem{selvaraju2017grad}
Ramprasaath~R Selvaraju, Michael Cogswell, Abhishek Das, Ramakrishna Vedantam,
  Devi Parikh, and Dhruv Batra.
\newblock Grad-cam: Visual explanations from deep networks via gradient-based
  localization.
\newblock In {\em Proceedings of the IEEE international conference on computer
  vision}, pages 618--626, 2017.

\end{thebibliography}






\end{document}